\documentclass[10pt,twocolumn,letterpaper]{article}

\usepackage[pagenumbers]{cvpr} 

\usepackage[colorlinks,linkcolor=red,citecolor=green]{hyperref}
\usepackage{url}
\usepackage{times}
\usepackage{epsfig}
\usepackage{wrapfig}
\usepackage{graphicx}
\usepackage{amsmath}
\usepackage{amssymb}
\usepackage{bm}
\usepackage{array}
\usepackage{booktabs}
\usepackage{multirow}
\usepackage{makecell}
\usepackage{footmisc}
\usepackage{footnote}
\usepackage{tablefootnote}
\newcommand{\eat}[1]{}

\usepackage[linesnumbered,boxed,ruled,commentsnumbered]{algorithm2e}

\DeclareMathOperator{\clip}{clip}
\newcommand{\norm}[1]{\left\lVert#1\right\rVert}

\DeclareMathOperator*{\argmin}{arg\,min}

\DeclareMathOperator{\sign}{sign}
\usepackage{algorithmic}

\usepackage[capitalize]{cleveref}
\crefname{section}{Sec.}{Secs.}
\Crefname{section}{Section}{Sections}
\Crefname{table}{Table}{Tables}
\crefname{table}{Tab.}{Tabs.}

\begin{document}

\title{Staircase Sign Method for Boosting Adversarial Attacks}
\author{Qilong Zhang\textsuperscript{1},
 Xiaosu Zhu\textsuperscript{1},
 Jingkuan Song\textsuperscript{1},
 Lianli Gao\textsuperscript{1},
 Heng Tao Shen\textsuperscript{1}\\
\textsuperscript{1}Center for Future Media and School of Computer Science and Engineering\\
University of Electronic Science and Technology of China, China\\
{\tt\small qilong.zhang@std.uestc.edu.cn,xiaosu.zhu@outlook.com,jingkuan.song@gmail.com}\\
{\tt\small lianli.gao@uestc.edu.cn,shenhengtao@hotmail.com}
}

\maketitle

\begin{abstract}
    Crafting adversarial examples for the transfer-based attack is challenging and remains a research hot spot. Currently, such attack methods are based on the hypothesis that the substitute model and the victim model learn similar decision boundaries, and they conventionally apply Sign Method (SM) to manipulate the gradient as the resultant perturbation. Although SM is efficient, it only extracts the sign of gradient units but ignores their value difference, which inevitably leads to a deviation. Therefore, we propose a novel Staircase Sign Method (S$^2$M) to alleviate this issue, thus boosting attacks. Technically, our method heuristically divides the gradient sign into several segments according to the values of the gradient units, and then assigns each segment with a staircase weight for better crafting adversarial perturbation. As a result, our adversarial examples perform better in both white-box and black-box manner without being more visible. Since S$^2$M just manipulates the resultant gradient, our method can be generally integrated into the family of FGSM algorithms, and the computational overhead is negligible. Extensive experiments on the ImageNet dataset demonstrate the effectiveness of our proposed methods, which significantly improve the transferability (\textit{i.e.}, on average, \textbf{5.1\%} for normally trained models and \textbf{12.8\%} for adversarially trained defenses). Our code is available at \url{https://github.com/qilong-zhang/Staircase-sign-method}.
\end{abstract}


\maketitle

\section{Introduction}
\begin{figure}[h]
		\centering
        \includegraphics[height=5.5cm]{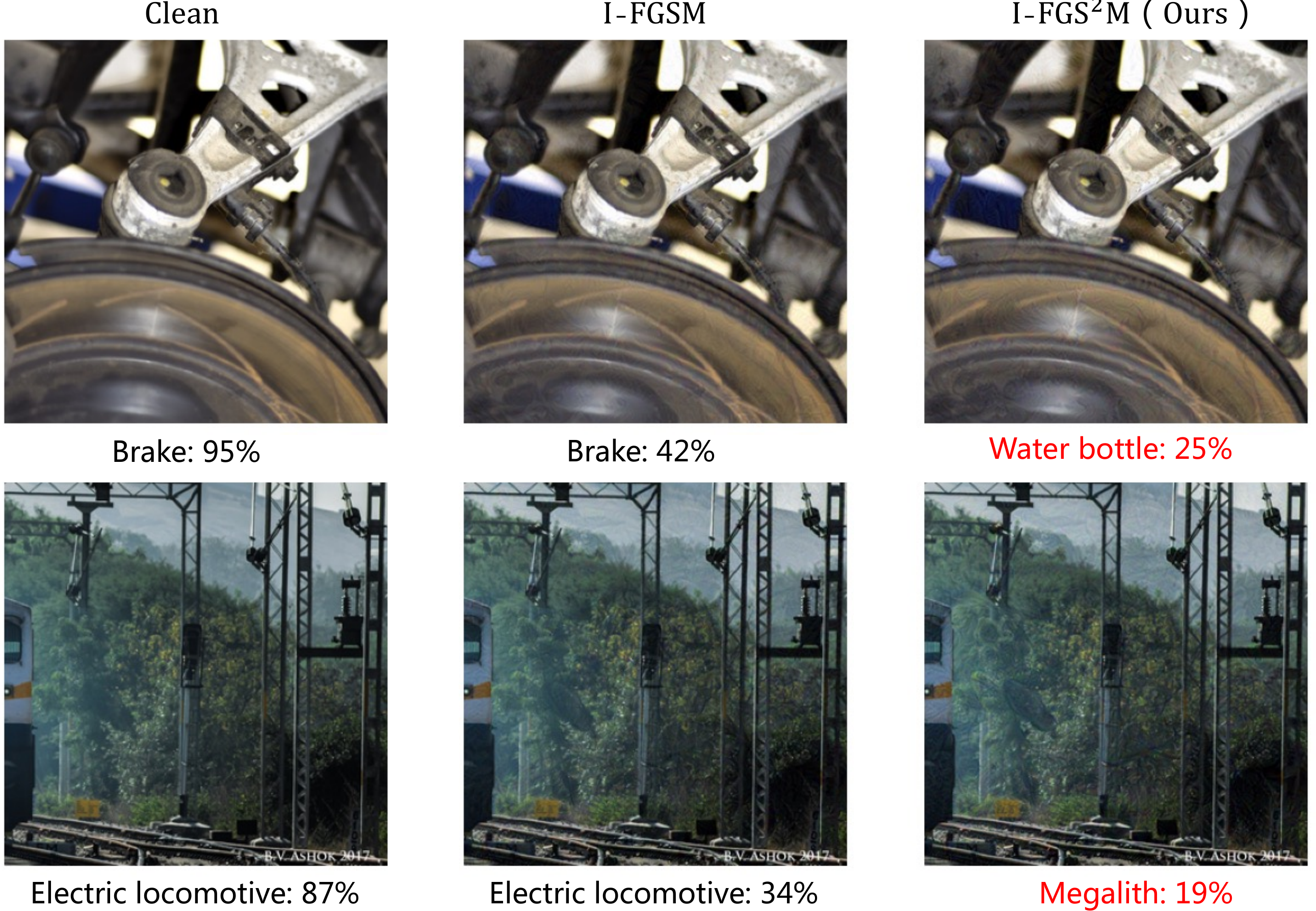}
	    \caption{We generate adversarial examples by I-FGSM and our proposed I-FGS$^2$M. Our method can successfully fool the victim model to classify resultant adversarial examples as the pre-set target label. 
	    }
		\label{show}
\end{figure}
With the remarkable performance of deep neural networks (DNNs) in various tasks, the robustness of DNNs are becoming a hot spot of the current research. However, DNNs are vulnerable to the adversarial examples~\cite{intriguing,fgsm,fgsm-,uap1,curlwhey} which are only added with relatively small perturbations but can fool state-of-the-art DNNs~\cite{inc-v3,inc-v4,res152,densenet} successfully. To make the matter worse, attacking in the physical world~\cite{Accessorize,ref_a13,T-Shirt,Zhang2022BIA,hit} is also practicable, which inevitably raises concerns in real-world applications such as self-driving cars.

To better evaluate the robustness of DNNs, various works have been proposed to seek the vulnerability of DNNs. Specifically, white-box attacks such as Deepfool~\cite{deepfool}, Carlini \& Wagner's (C\&W) method~\cite{c&w} and Adaptive Auto Attack (A$^3$)~\cite{a3} can achieve impressive performance with the complete knowledge of the victim model (\textit{a.k.a.} black-box model), \textit{e.g.}, gradient and structure. 
However, deployed DNNs are usually transparent to unauthorized users for security, and thus the adversary cannot base on any knowledge of the victim model. 
Therefore, resorting to cross-model transferability~\cite{intriguing,uap,Nasser2019cross,admix,gao2021feature,fia,Naseer2021on} of adversarial examples is a common practice. That is to say, the adversarial examples crafted via known white-box models (\textit{a.k.a.} substitute model) are also dangerous for other unknown black-box models, which makes the black-box transfer-based attack possible. In this field, Goodfellow \textit{et al.}~\cite{fgsm} hypothesize that the vulnerability of DNNs is their linear nature. 
Since raw gradient magnitude is extremely small (\textit{e.g.} the minimal unit in gradient $\approx 10^{-9}$) and digital images usually use 8 bits per pixel, they propose Fast Gradient Sign Method (FGSM)~\cite{fgsm} so that each pixel can be fully perturbed with only a single step. 
Conventionally, the following transfer-based iterative attack methods~\cite{mifgsm,difgsm,sinifgsm,pifgsm,sgm} are all based on Sign Method (SM) to boost adversarial attack.

\begin{figure*}
	\centering
	\includegraphics[height=4.25cm]{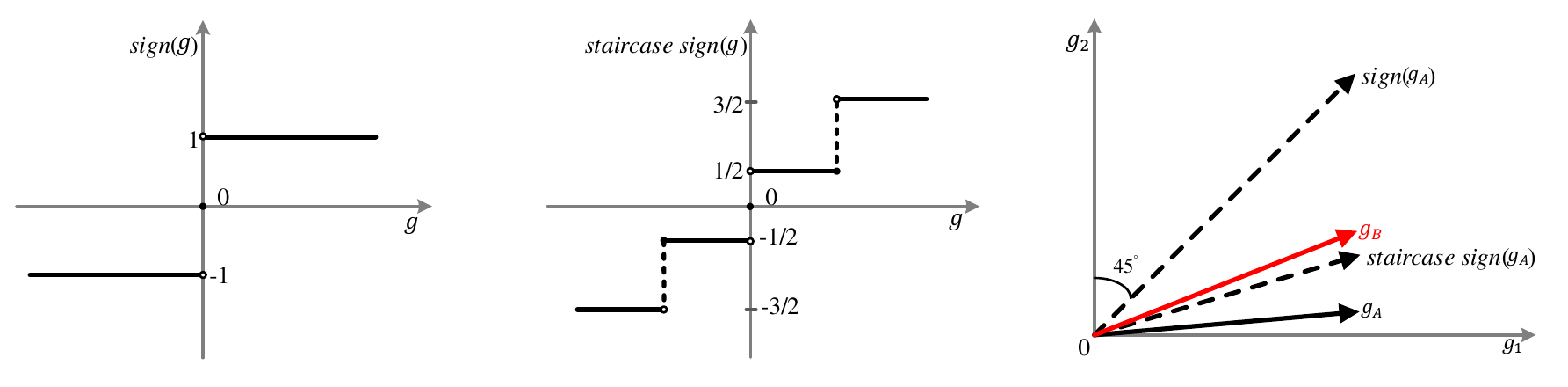}
	\caption{
		Sign function vs. Staircase sign function (take $K=2$ for example). \textbf{Left \& Middle:} the graph of functions. \textbf{Right:} illustration of the gradient direction of the substitute model $g_A$, black-box model $g_B$, and the resultant update directions of sign and staircase sign function w.r.t. $g_A$.
		For sign function, each unit of the resultant direction is the same.
		By contrast, our method reflects the difference between units, and is more close to both $g_A$ and $g_B$.}
	\label{function}
\end{figure*}
However, there is a limitation in SM, \textit{i.e.}, ignores the difference among each unit in the gradient vector. 
As illustrated in Figure~\ref{function}(a), the update direction obtained by the sign function is that whether the partial derivative of loss function at each pixel is positive, negative or zero. Since the transferability phenomenon is mainly due to the fact that decision boundaries around the same data point of different models are similar, a naive application of SM results in a poor gradient estimation (as depicted in Figure~\ref{function}(c)). 
Consequently, the adversarial examples especially for targeted ones may deviate from the \textit{global optimal attack region} where both the substitute model and the victim model can be fooled, thus decreasing the transferability. 

Motivated by this,  we propose a \textbf{Staircase Sign Method (S$^2$M)} to effectively utilize the gradient for the substitute model, thus more closely approximating the gradients of both the black-box and white-box models. Technically, our proposed method first utilizes the sign function to roughly get the gradient direction for the substitute model, then heuristically assigns different weights for each pixel by our staircase sign function. In short, we merely manipulate the sign perturbation added on the image. Thus, S$^2$M can be generally integrated into the family of FGSM algorithms. Based on I-FGSM~\cite{ifgsm}, we propose its variant I-FGS$^2$M (Algorithm \ref{A1}) which can also serve as an iterative attack baseline to be combined with existing approaches, \textit{e.g.}, Input diversity~\cite{difgsm}, Poincar{\'e} space loss~\cite{pom}, and Patch-wise++ method~\cite{pifgsm++}. To sum up, our main contributions are as follows:
\begin{itemize}
	\item To the best of our knowledge, we are the first to empirically and experimentally point out the poor gradient estimation limitation of Sign Method (SM) in transfer-based attacks, which causes the adversarial examples to deviate from the \textit{global optimal attack region}.
	
	\item We propose a novel Staircase Sign Method (S$^2$M) to alleviate this problem. Notably, our method is simple but effective, and can be integrated into the family of FGSM algorithms. 
	
	\item Extensive experiments on the ImageNet dataset~\cite{imagenet} demonstrate the effectiveness of our proposed attacks which consistently outperform vanilla FGSM-based non-targeted \& targeted ones in both black-box and white-box manner.  
\end{itemize}

\section{Related Works}
\subsection{Transfer-based Black-box Attacks}
\label{attack methods}
Unlike white-box attacks, black-box attacks cannot obtain the gradient or parameters of the victim model. Although query-based black-box attacks~\cite{zoo,query2,sa} can be applied in this manner, a large number of queries is computationally expensive. Thus, we resort to the transferability of adversarial examples in this paper.

For non-targeted attacks, 
Goodfellow \textit{et al.}~\cite{fgsm} quantify the gradient by the sign function and propose single-step FGSM with the step size equal to maximum perturbation. However, perturbing images with single-step attacks usually cannot get a high success rate on the white-box model. Therefore, Kurakin \textit{et al.}~\cite{ifgsm} propose I-FGSM which applies FGSM multiple times with a small step size. Considering that iterative methods usually sacrifice transferability to improve the white-box performance, Dong \textit{et al.}~\cite{mifgsm} integrate momentum term into the iterative process to avoid adversarial examples falling into local optimum. Xie \textit{et al.}~\cite{difgsm} apply random transformations to the input images to alleviate the overfitting problem. Wu \textit{et al.}~\cite{sgm} explore the security weakness of skip connections~\cite{res152} to boost adversarial attacks. 
To effectively evade defenses, Dong \textit{et al.}~\cite{tifgsm} propose a translation-invariant attack method to smooth the perturbation. Lin \textit{et al.}~\cite{sinifgsm} adapt Nesterov accelerated gradient and leverage scale-invariant property of DNNs to optimize the perturbations. Nasser \textit{et al.}~\cite{Nasser2019cross} train cross-domain generators by using different training data distribution.
Gao \textit{et al.}~\cite{pifgsm} craft patch-wise noise to further increase the success rate of adversarial examples. 
Zhang \textit{et al.}~\cite{Zhang2022BIA}  propose a Beyond ImageNet Attack, \textit{i.e.}, with only the knowledge of the ImageNet domain, to investigate the transferability towards black-box domains.

However, targeted attacks are more challenging which need to guide the victim model to predict a specific target class with high confidence rather than just cause misclassification. In this setting, Li \textit{et al.}~\cite{pom} replace the cross-entropy loss with Poincar{\'e} distance and introduce triple loss to make adversarial examples close to the target label. Gao \textit{et al.}~\cite{pifgsm++} extend non-targeted \cite{pifgsm} to targeted version and adopt temperature term to push the adversarial examples into the \textit{global optimal  attack region} of DNNs. Instead of optimizing the output distribution with a few iterations, Zhao \textit{et al.}~\cite{secondlook} directly maximize the target logits with more iterations.
To make the adversarial examples more transferable, 
several researchers~\cite{feature14,featurespace} turn to directly optimize intermediate features instead of output distribution. Naseer \textit{et al.}~\cite{Naseer2021on} train generators to match ``distribution" of the target class.
Although \cite{fda1,fda2,Naseer2021on} have achieved impressive performance in this way, both training specific auxiliary models and generators for each target class is time-consuming. Therefore, we resort to FGSM-based attacks in this paper.

\subsection{Defense Methods}
\label{defense methods}
With the development of adversarial examples, researchers pay considerable attention to the robustness of DNNs, and various defense methods are proposed to circumvent potential risks. Guo \textit{et al.}~\cite{guo} apply multiple input transformations such as JPEG compression~\cite{jpeg}, total variance minimization~\cite{tvm} and image quilting~\cite{imagequilt} to recover from adversarial perturbations. 
Theagarajan \textit{et al.}~\cite{shieldnets} introduce probabilistic adversarial robustness to neutralize adversarial attacks by concentrating sample probability to adversarial-free zones. Liao \textit{et al.}~\cite{HGDarxiv} propose a high-level representation guided denoiser (HGD) to suppress adversarial perturbations. Xie \textit{et al.}~\cite{RP} mitigate adversarial effects through random resizing and padding (R\&P).

Although the above methods are efficient, \textit{i.e.}, do not require a time-consuming training process, the adversarial training defense mechanism is more robust in practice. In this field, Madry \textit{et al.}~\cite{mary} 
adopt a natural saddle point formulation to cover the blind spots of DNNs.
Tram{\`{e}}r \textit{et al.}~\cite{eat} 
introduce ensemble adversarial training which augments training data with perturbations transferred from other models. Xie \textit{et al.}~\cite{featuredenoising} impose constraints at the feature level by denoising technique. Naseer \textit{et al.}~\cite{NRP} design a Neural Representation Purifier (NRP) model that learns to clean adversarial perturbed images based on the automatically derived supervision.

\section{Methodology}
\label{methodology}
\subsection{Problem Formulation}
Before introducing our algorithm in detail, we first describe the background knowledge of generating adversarial examples. Given a DNN network $f\left(\bm{x}\right):\bm{x}\in\mathcal{\bm{X}}\longrightarrow y\in\mathcal{Y}$, it takes an input $\bm{x}$ (\textit{e.g.}, a clean image) to return its true label $y$. For targeted attacks\footnote{Non-targeted attacks are discussed in Appendix Sec.3}, it requires us to find a relatively small perturbation $\bm{\delta}$ to
satisfy $f\left(\bm{x^{*}}\right)=y^{*}$, where $\bm{x^{*}}=\bm{x}+\bm{\delta}$ is the generated adversarial example and $y^{*}$ is the preset target label.

To make the resultant adversarial examples are perceptually indistinguishable from the clean ones, the adversary usually sets a small perturbation upper bound $\epsilon$, and lets $\norm{\bm{\delta}}_\infty \leq \epsilon$. 
By minimizing the loss function $\operatorname{J}\left(\bm{x}^*, y^*\right)$, \textit{e.g.}, cross entropy loss, the constrained optimization problem can be denoted as:
\begin{equation}
	\argmin_{x^{*}\in\mathcal{\bm{X}}}{\operatorname{J}\left(\bm{x^{*}}, y^{*}\right)},\;\; \mathit{s.t.}\; \norm{\bm{x^{*}}-\bm{x}}_\infty \leq \epsilon.
\end{equation}

For targeted attacks (\textit{e.g.}, take basic I-FGSM as an examples) the resultant adversarial example at iteration $t+1$ can be formally written as:
\begin{equation}
	\bm{x^{*}_{t+1}} = \clip_{\bm{x}, \epsilon}\{\bm{x^{*}_{t}} - \alpha \cdot \sign\left(\nabla_{\bm{x^{*}_t}} \operatorname{J}\left(\bm{x^{*}_t}, y^{*}\right)\right)\},
	\label{I-FGSM}
\end{equation}
where $\clip_{\bm{x}, \epsilon}\left(\cdot\right)$ keeps the adversarial example $\bm{x}^{*}$ within the $\epsilon$-ball of $\bm{x}$, and $\alpha$ is the step size.


\subsection{Rethinking the Sign Method}
\label{motivation}
Recently, FGSM-based algorithms plays a key role in the field of transferability. Particularly, these attacks are all based on SM to generate adversarial examples. In addition to the linear hypothesis~\cite{fgsm}, the motivation of SM
is to modify more information for each pixel than directly adding the gradient, especially for single-step attacks. Besides, manipulating the gradients by SM for iterative attacks can quickly reach the boundary of $\ell_\infty$-ball with only a few iterations~\cite{mifgsm,tifgsm}. 

\eat{
\begin{table}[htbp]
	\centering
	\resizebox{0.95\linewidth}{!}{
		\begin{tabular}{cccccc}
			\hline
			& Hold-out $\uparrow$ & Ensemble $\uparrow$ & AoE~\cite{pifgsm++} $\uparrow$ & RMSE $\downarrow$ \\
			\hline
			\hline
			Random-SM & 0.0    & 0.0   & 0.0 & 15.61 \\ 
			\hline
			I-FGSM~\cite{ifgsm} & 1.2    & 99.9   & 94.7 & 4.23  \\
			\hline
			MI-FGSM~\cite{mifgsm} & 6.3    & 99.9   & 94.5 & 11.18 \\
			\hline
			DI$^2$-FGSM~\cite{difgsm} & 15.4   & 91.8   & 77.8 & \textbf{4.17}  \\
			\hline
			TI-FGSM\tablefootnote{For consistency, here we term TI-BIM~\cite{tifgsm} as TI-FGSM.}~\cite{tifgsm} & 1.6    & 99.9   & 94.0 & 4.27  \\
			\hline
			Po-FGSM~\cite{pom} & 1.1    & \textbf{100.0}  & 88.7 & 4.20  \\
			\hline
			PI-FGSM~\cite{pifgsm} & \textbf{22.4}   & 99.9 & \textbf{98.1} & 13.87  \\
			\hline
	\end{tabular}}%
	\caption{The average success rates (\%) of targeted attacks and RMSE under the same $\ell_\infty$ norm constraint (\textit{i.e.} 16) on 1,000 images. The adversarial examples are crafted via an ensemble of Inc-v4~\cite{inc-v4}, IncRes-v2~\cite{inc-v4} and Res-152~\cite{res152}, and the hold-out model is Inc-v3~\cite{inc-v3}. For Random-SM, we simply sample noise from the Gaussian distribution, then manipulated by SM.}
	\label{tab:rethink}%
\end{table}%
}
\eat{
Intuitively, SM is useful for improving transferability because a bigger perturbation makes the feature of adversarial example close to that of the target class image more easily.
However, relying on the large RMSE alone is insufficient.
As demonstrated in Table~\ref{tab:rethink}, Random-SM, which is not guided by the gradient of the substitute model, cannot successfully attack any models but with the highest RMSE. Besides, DI$^2$-FGSM remarkably outperforms MI-FGSM by 9.1\% (Hold-out) but only with 37.3\% RMSE of MI-FGSM. In other words, resorting to a more accurate gradient estimation for the black-box model in a few iterations is crucial. 
}

However, the transferability of adversarial examples is mainly based on the phenomenon that 
decision boundaries of different models are similar. Since targeted attacks need to guide the adversarial examples into a specific territory of the target class, directly applying SM inevitably discards significant information of the gradient of the substitute model. 
As the example shown in Figure~\ref{function}, the direction derived from the SM significantly deviates from the gradient of the victim model. Consequently, the resultant perturbation deviates from the target territory, thus decreases the transferability. 


\subsection{Staircase Sign Method}
\label{attack algorithem}
Motivated by the limitation of SM, we propose a novel \textbf{Staircase Sign Method (S$^2$M)} to alleviate this problem. Figure~\ref{function} depicts the difference between sign function and our proposed staircase sign function. Since our method merely manipulates the gradient at each iteration, it can be generally integrated into the family of FGSM algorithms. For simplicity purpose, we only take our variant I-FGS$^2$M (summarized in Algorithm~\ref{A1}) as an example to show the integration process.

Technically, our method can be mainly divided into four steps. Firstly, as with the other methods, \textit{e.g.},~\cite{tifgsm,difgsm,pifgsm,fia,admix}, we need to compute the gradient $\bm{G}_t$ at $t$-iteration of the substitute model with respect to the input (in line 5):
\begin{equation}
	\bm{G}_t = \nabla_{\bm{x^{*}_t}} \operatorname{J}\left( \bm{x^{*}_t}, y^{*}\right)
\end{equation}
Secondly, we calculate the $p-$th percentile $g^p$ of $\lvert\bm{G}_t\rvert$ (in line 7) according to the number of staircase $K$, where $p$ ranges from $100/K$ to $100$ with the percentile interval $\tau=100/K$. 
Thirdly, we assign the staircase weights $\bm{W}_t$ according to $g^p_t$ by Eq.~(\ref{W}) (in line 8):
\begin{equation}
	\label{W}
	\bm{W}_t=
	\begin{cases}
		\frac{\tau}{100},&  g^{0}_t \leq \lvert\bm{G}^{i,j}_t\rvert \leq g^{\tau}_t, \\
		\frac{3\tau}{100},&  g^{\tau}_t < \lvert\bm{G}^{i,j}_t\rvert \leq g^{2\tau}_t, \\
		&\vdots  \\
		\frac{\left(2k+1\right)\tau}{100},&  g^{p-\tau}_t < \lvert\bm{G}^{i,j}_t\rvert \leq g^p_t, \\
		&\vdots  \\
		\frac{\left(2K-1\right)\tau}{100},&  g^{100-\tau}_t < \lvert\bm{G}^{i,j}_t\rvert \leq g^{100}_t.
	\end{cases} \\
\end{equation}	
where $k$ ranges from 0 to $K-1$, and also equals to $p/\tau-1$. As a result, our $\bm{W_t}$ is bounded in $\left[1/K, 2-1/K\right] \subseteq \left[0,2\right]$. Finally, combined with the sign direction of $\bm{G}_t$, we rewrite Eq.~\ref{I-FGSM} to craft our adversarial examples $\bm{x}^*_t$ (in lines 11): 
\begin{equation}
	\bm{x^{*}_{t+1}} = \clip_{{\bm{x}}, \epsilon}\{ \bm{x^{*}_{t}} - \alpha\cdot \sign\left(\bm{G}_t\right) \odot \bm{W}_t\},	
\end{equation}	
where $\odot$ is Hadamard product.

\textbf{Proposition 1} \textit{Assume that $\bm{W}_t$ is i.i.d. (subject to the uniform distribution) for all $t\in\left[0,T-1\right]$ , the adversarial examples can reach the boundary of $\varepsilon$-ball.}

\begin{figure}[t]
	\centering
	\includegraphics[height=3cm]{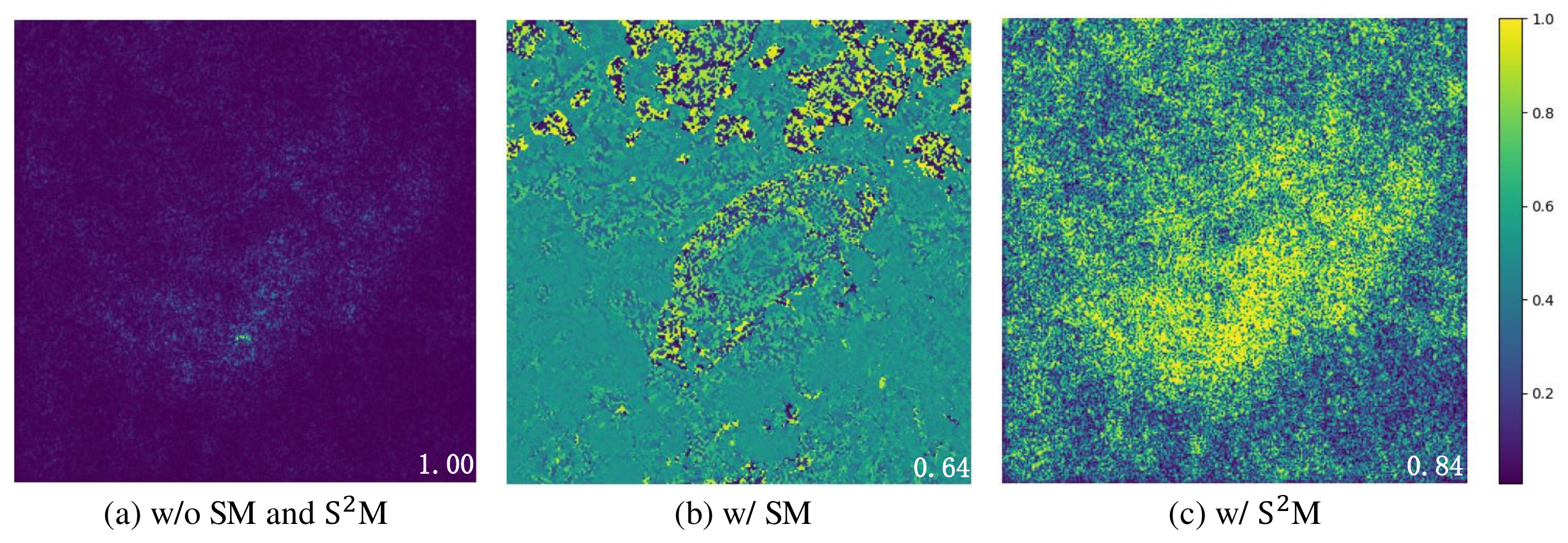}
	\caption{We visualize the perturbations of an image at the first iteration. All perturbations are normalized to [0, 1]. \textbf{(a)}: the gradient of an ensemble of Inc-v4~\cite{inc-v4}, IncRes-v2~\cite{inc-v4} and Res-152~\cite{res152}; \textbf{(b)\&(c)}: the results of SM and S$^2$M ($K=64$) w.r.t. (a).
		The numbers in the lower right corner indicate the average cosine similarity between (a) and the other two perturbations on 1,000 images. Notably, our S$^2$M not only keeps the perturbation magnitude but also has higher cosine similarity to (a).}
	\label{heatmap}
\end{figure}

\textit{Proof.} 
Due to the fact that $\norm{\cdot}_\infty=\max\left(\operatorname{abs}\left(\cdot\right)\right)$, here we only discuss $\bm{W_t}^{i,j}$ (the element of the i-th row and j-th column of $\bm{W_t}$) which is also i.i.d. Therefore, 
\begin{equation}
\begin{aligned}
    \norm{\sum_{t = 0}^{T-1}\alpha\cdot \sign\left(\bm{G_t}\right) \odot \bm{W_t}}_\infty&=\norm{\sum_{t = 0}^{T-1}\alpha\cdot\bm{W}_t^{i,j}}_\infty\\&=\alpha\cdot\sum_{t = 0}^{T-1}\bm{W}_t^{i,j}.
\end{aligned}
\end{equation}
Besides, 
\begin{equation}
\begin{aligned}
    \mathbb{E}(\bm{W}_t^{i,j})&=\sum_{k = 0}^{K-1}\frac{1}{K}\cdot\frac{(2k+1)\tau}{100}\\&=\frac{1}{100K}\cdot\left(\tau+3\tau+...+\left(2K-1\right)\tau\right)\\&=\frac{K\tau}{100}=1
\end{aligned}
\end{equation}
So we have:
\begin{equation}
\begin{aligned}
\mathbb{E}\left(\norm{\sum\limits_{t = 0}^{T-1}\alpha\cdot \sign\left(\bm{G}_t\right) \odot \bm{W}_t}_\infty\right)
&=\alpha\cdot\sum\limits_{t = 0}^{T-1}\mathbb{E}\left(\bm{W}^{i,j}_t\right)\\
&=\alpha\cdot T \\
&=\varepsilon
\end{aligned}
\end{equation}
In fact, S$^2$M is equivalent to applying adaptive weight for each pixel in sign noise. With the help of our S$^2$M, the poor gradient estimation problem caused by SM can be effectively alleviated. 
As demonstrated in Figure~\ref{heatmap}, the cosine similarity between the gradients (a) and the perturbations manipulated by our proposed S$^2$M (c) is up to \textbf{0.84}, while the result of SM is only 0.64. Please note that our S$^2$M does not aim to make the cosine similarity close to 1.0. This is because the victim model is only similar to the substitute model, but it cannot be exactly the same. As demonstrated in Figure~\ref{function}(c), ``overfitting" on the substitute model will also enlarge the gap with the victim model.

The adversarial examples are shown in Figure~\ref{show}. Compared with I-FGSM which cannot effectively decrease the confidence of true class, our proposed variant successfully misleads the model to classify our resultant adversarial examples as the pre-set target classes.
\subsection{Attacking an Ensemble of Models}
\label{ensemble}
To craft adversarial examples with high transferability, attacking an ensemble of models~\cite{delving,mifgsm} is an effective strategy, especially for black-box attacks. It is mainly because crafting adversarial examples on multiple models has the potential to capture the \textit{global optimal attack region} easily. In this paper, we follow the ensemble strategy of~\cite{mifgsm}, which fuses the logits (the output before the softmax) of an ensemble of $M$ models:
\begin{equation}
	l\left(\bm{x}\right)=\sum\limits_{m=1}^{M}u_m l_m\left(\bm{x}\right),
\end{equation}
where $l_m\left(\cdot\right)$ is the logits of $\mathit{m}$-th model, and $u_m$ is its ensemble weight with $u_m \textgreater 0$ and $\sum_{m=1}^M u_m=1$.

\begin{algorithm}[h]
\caption{I-FGS$^2$M}
\SetAlgoNoLine 
\SetKwInOut{Input}{\textbf{Input}}\SetKwInOut{Output}{\textbf{Output}}
\Input{The cross-entropy loss function $\operatorname{J}$ of our substitute model; iterations $T$; $\ell_{\infty}$ constraint $\epsilon$; a clean image $\bm{x}$ (Normalized to $\left[-1, 1\right]$) and its corresponding true label $y$; the target label $y^*$; the number of staircase $K \left(\ge 2\right)$; 
}
\Output{The adversarial example $\bm{x}^{*}$;
}
\begin{algorithmic}[1]
\STATE	$\bm{x^{*}_0} = \bm{x}$;\\ 
\STATE	$\alpha=\epsilon/T$; $\tau=100/K$;\\
\STATE  Initialize staircase weights $\bm{W}$ to 0, and $p$ to $100/K$;
\FOR{$t \leftarrow 0$ \KwTo $T$}
\STATE $\bm{G}_t$ = $\nabla_{\bm{x^{*}_t}} \operatorname{J}\left( \bm{x^{*}_t}, y^{*}\right)$;\quad\\

\FOR{$k \leftarrow 0$ \KwTo $K$}
\STATE calculate the $p$-th percentile $g^p_t$ of $\lvert\bm{G}_t\rvert$;\\
\STATE      $
			\bm{W}_t\!=\!
			\begin{cases}
				\frac{\tau}{100},\!&  g^{0}_t \!\leq\! \lvert\bm{G}_t^{i,j}\rvert \!\leq\! g^{\tau}_t, \\
				\frac{3\tau}{100},\!&  g^{\tau}_t \!<\! \lvert\bm{G}_t^{i,j}\rvert \!\leq\! g^{2\tau}_t, \\
				&\vdots  \\
				\frac{(2k+1)\tau}{100}\!,\!&  g^{p-\tau}_t \!<\! \lvert\bm{G}_t^{i,j}\rvert \!\leq\! g^p_t, \\
				&\vdots  \\
				\frac{(2K-1)\tau}{100}\!,\!&  g^{100-\tau}_t \!<\! \lvert\bm{G}_t^{i,j}\rvert \!\leq\! g^{100}_t.
			\end{cases} 
			$ \\
\STATE 		$p = p + \tau$ 
\ENDFOR
\STATE	$\bm{x^{*}_{t+1}} = \clip_{{\bm{x}}, \epsilon}\{ \bm{x^{*}_{t}} - \alpha\cdot \sign\left(\bm{G}_t\right) \odot \bm{W}_t\}$; 	\quad\\
\STATE		$\bm{x^{*}_{t+1}} = \clip\left(\bm{x^{*}_{t+1}}, -1, 1\right)$; 
\ENDFOR
\STATE	Return $\bm{x^{*}} = \bm{x^{*}_{T}}$;
\end{algorithmic}
\label{A1}
\end{algorithm}

\section{Experiments}
To demonstrate the effectiveness of our staircase sign mechanism, we conduct extensive experiments based on the family of FGSM methods. Firstly, we introduce the setup of experiments in Sec.~\ref{setup}. Secondly, we analyze the effect of staircase number in Sec.~\ref{exp:k}. Then the attack success rates for normally trained models, black-box robust models and white-box robust models are reported in Sec.~\ref{exp:nt}, Sec.~\ref{exp:df} and Sec.~\ref{exp:wdf}, respectively. After that, we discuss the effect of different $\epsilon$ in Sec.~\ref{exp:epsilon}.
Finally, we give an insight into our method in Sec.~\ref{insight}.
Due to the space limitation, non-targeted attacks are discussed in Appendix Sec.3. Notably, our non-targeted FGS$^2$M variants can outperform vanilla FGSM ones by \textbf{19.1\%} at most. 

\begin{figure*}[t]
	\centering
	\includegraphics[height=4.75cm]{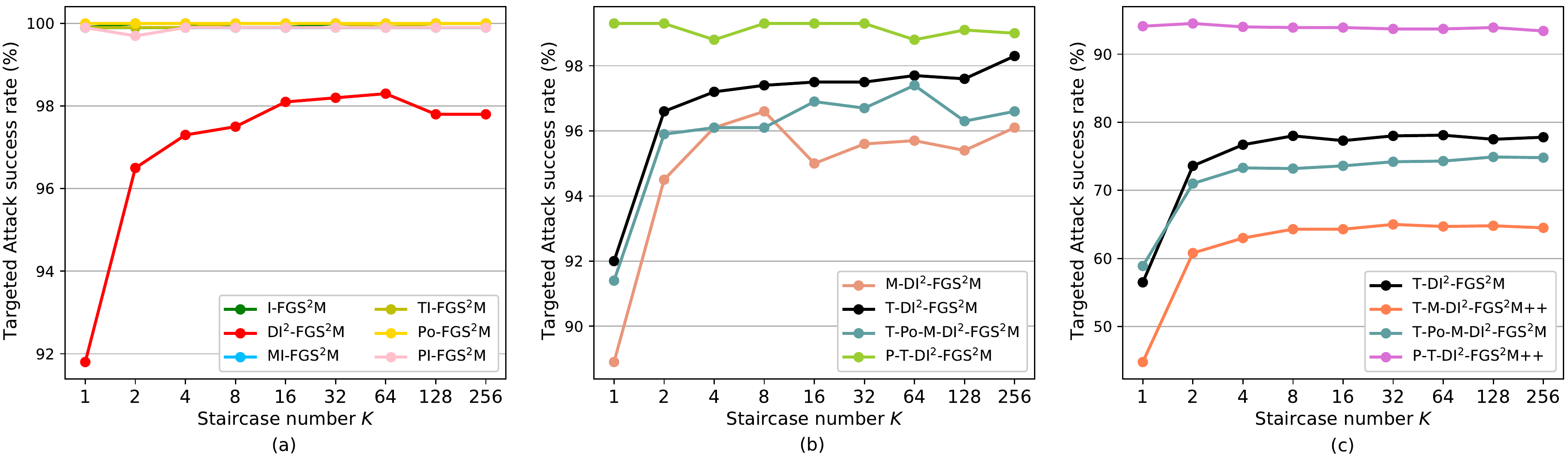}
	\caption{The success rates (\%) of targeted white-box attack (Ensemble) for different methods w.r.t. staircase number $K$ ($K\!=\!1$ denotes SM and $K\!\ge\!2$ denotes S$^2$M). 
		For (a) and (b), the adversarial examples are crafted via an ensemble of Inc-v4, IncRes-v2 and Res-152. For (c), the white-box models are an ensemble of Inc-v3, Inc-v4, IncRes-v2, Res-152, Res-101, Res-50, Inc-v3$_{ens4}$ and IncRes-v2$_{ens}$.}
	\label{fig:ensemble}
\end{figure*}
\begin{figure*}[t]
	\centering
	\includegraphics[height=4.75cm]{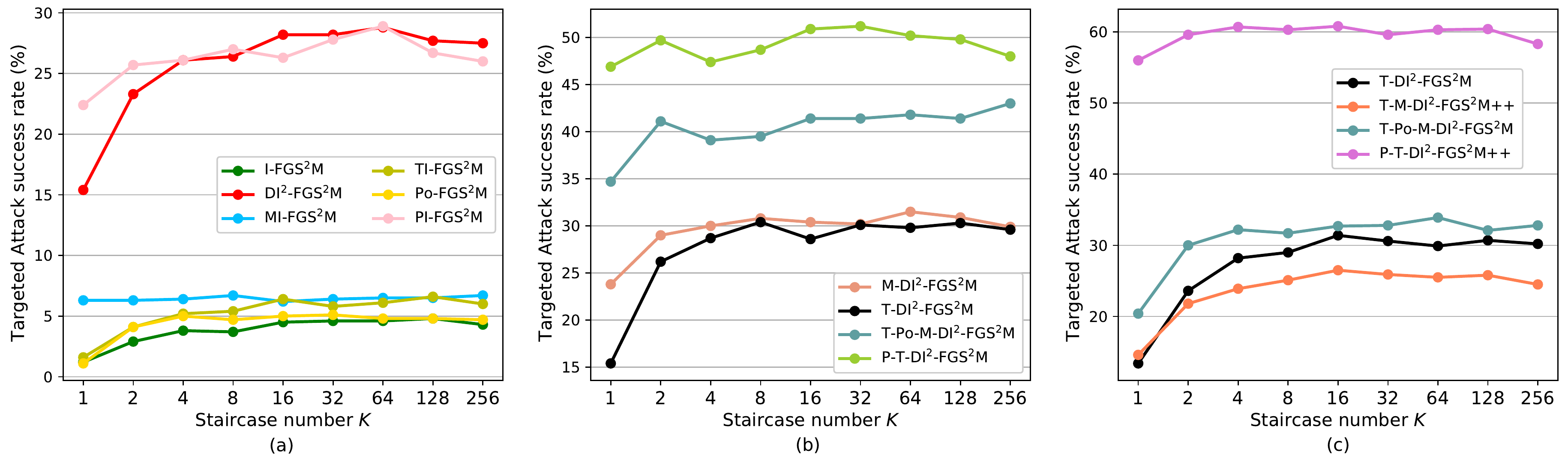}
	\caption{The success rates (\%) of targeted black-box attack (Hold-out) for different methods w.r.t. staircase number $K$ ($K\!=\!1$ denotes SM and $K\!\ge\!2$ denotes S$^2$M). 
		For (a) and (b), the adversarial examples are crafted via an ensemble of Inc-v4, IncRes-v2 and Res-152, and the hold-out model is Inc-v3. For (c), the white-box models are an ensemble of Inc-v3, Inc-v4, IncRes-v2, Res-152, Res-101, Res-50, Inc-v3$_{ens4}$ and IncRes-v2$_{ens}$, and the hold-out model is Inc-v3$_{ens3}$.}
	\label{fig:hold}
\end{figure*}
\subsection{Setup}
\label{setup}
\textbf{Networks}: In this paper, we consider sixteen well-known models, including six normally trained models: Inception-v3 (Inc-v3)~\cite{inc-v3}, Inception V4 (Inc-v4) \cite{inc-v4}, Inception-ResNet V2 (IncRes-v2)~\cite{inc-v4}, ResNet-50 (Res-50), ResNet-101 (Res-101) and ResNet-152 (Res-152) \cite{res152}, three ensemble adversarial training models: Inc-v3$_{ens3}$, Inc-v3$_{ens4}$ and IncRes-v2$_{ens}$~\cite{eat}, another four widely used defenses\footnote{For HGD, R\&P, NIPS-r3, we adopt the official models provided in corresponding papers. NRP uses Inc-v3$_{ens3}$ to classify purified images.}: HGD~\cite{HGDarxiv}, NIPS-r3~\cite{nips_r3}, R\&P~\cite{RP}, and NRP~\cite{NRP}, and three white-box robust feature denosing models: Res152$_B$, Res152$_D$ and ResNext101$_{DA}$~\cite{featuredenoising}.


\textbf{Dataset}: We conduct our experiments on ImageNet-compatible dataset\footnote{\url{https://github.com/tensorflow/cleverhans/tree/master/examples/nips17_adversarial_competition/dataset}}. This dataset is comprised of 1,000 images, and widely used in recent FGSM-based attacks~\cite{mifgsm,difgsm,tifgsm,pifgsm,pom,pifgsm++}. The target label for each image in this dataset is pre-set and usually different.

\textbf{Parameters}: To comprehensively compare the performance between different attack methods, in our experiments, the maximum perturbation $\epsilon$ is set to 16 unless otherwise stated, 
the iteration $T$ of all methods is 20, and thus the step size $\alpha=\epsilon/T=0.8$.
When attacking an ensemble of $M$ models simultaneously, the weight for the logits of each model is equal, \textit{i.e.}, $1/M$. For MI-FGSM~\cite{mifgsm}, the decay factor $\mu=1.0$. For DI-FGSM~\cite{difgsm}, the transformation probability $p=0.7$. For TI-FGSM~\cite{tifgsm}, when the victim model is in normally trained models, the Gaussian kernel length is $5\times5$, and $15\times15$ for defense models. For Po-FGSM~\cite{pom}, we set $\lambda=0.1$. For PI-FGSM~\cite{pifgsm} and PI-FGSM++~\cite{pifgsm++}, the project kernel length is $3\times 3$, the amplification factor $\beta=10$, and the project factor $\gamma=0.8\beta\alpha$ unless otherwise stated. The temperature $\tau$ for PI-FGSM++ is set to 1.5. For our S$^2$M, the number of staircase $K$ is set to 64. Please note that the parameters of each method are fixed no matter what methods are combined.

\textbf{Evaluation Metrics}: We use attack success rate (\%) to evaluate the performance of targeted attacks. In particular, ``Hold-out" is the success rate of the black-box models (\textit{i.e.} transferability), ``Ensemble" denotes the white-box success rates for an ensemble of models, and ``AoE"~\cite{pifgsm++} averages the white-box success rate of each model.

\begin{table*}[h]
	\centering
		\caption{The success rates (\%) of targeted FGSM-based/FGS$^2$M-based attacks. We study four models---Inc-v3, Inc-v4, Res-152 and IncRes-v2, and adversarial examples are crafted via an ensemble of three of them. In each column, ``-" denote the hold-out model. 
		}
	\resizebox{0.85\linewidth}{!}{
		\begin{tabular}{c|c|c|c|c|c|c}
			\hline
			Metrics &  Attacks     & -Inc-v3 & -Inc-v4 & -Res-152 & -IncRes & AVG. \\
            \hline
            \hline
			\multirow{10}[4]{*}{\makecell[c]{Ensemble}} & I~--~(FGSM~/~FGS$^2$M)     & 99.9~/~\textbf{100.0} & 99.9~/~\textbf{100.0} & \textbf{100.0}~/~\textbf{100.0} & \textbf{100.0}~/~\textbf{100.0} & \textbf{100.0}~/~\textbf{100.0} \\
			& MI~--~(FGSM~/~FGS$^2$M)    & \textbf{99.9}~/~\textbf{99.9} & 99.9~/~\textbf{100.0} & \textbf{100.0}~/~\textbf{100.0} & \textbf{100.0}~/~\textbf{100.0} & \textbf{100.0}~/~\textbf{100.0} \\
			& ~~DI$^2$~--~(FGSM~/~FGS$^2$M)   & 91.8~/~\textbf{98.3} & 93.3~/~\textbf{98.2} & 94.5~/~\textbf{98.7} & 94.8~/~\textbf{98.5} & 93.6~/~\textbf{98.4} \\
			& TI~--~(FGSM~/~FGS$^2$M)    & \textbf{99.9}~/~\textbf{99.9} & 99.8~/~\textbf{99.9} & 97.0~/~\textbf{99.9} & \textbf{100.0}~/~\textbf{100.0} & 99.2~/~\textbf{99.9} \\
			& Po~--~(FGSM~/~FGS$^2$M)    & \textbf{100.0}~/~\textbf{100.0} & 99.9~/~\textbf{100.0} & \textbf{100.0}~/~\textbf{100.0} & \textbf{100.0}~/~\textbf{100.0} & \textbf{100.0}~/~\textbf{100.0} \\
			& PI~--~(FGSM~/~FGS$^2$M)    & \textbf{99.9}~/~99.7 & \textbf{99.9}~/~\textbf{99.9} & \textbf{100.0}~/~99.9 & \textbf{99.8}~/~\textbf{99.8} & \textbf{99.9}~/~99.8 \\
			\cline{2-7}         
			& M-DI$^2$-FGSM~--~(FGSM~/~FGS$^2$M)  & 88.9~/~\textbf{95.7} & 90.4~/~\textbf{96.7} & 91.0~/~\textbf{96.2} & 92.8~/~\textbf{98.3} & 90.8~/~\textbf{96.7} \\
			& T-DI$^2$-FGSM~--~(FGSM~/~FGS$^2$M)  & 92.0~/~\textbf{97.7} & 92.2~/~\textbf{97.7} & 93.2~/~\textbf{98.2} & 94.0~/~\textbf{98.5} & 92.9~/~\textbf{98.0} \\
			& T-Po-M-DI$^2$~--~(FGSM~/~FGS$^2$M)  & 91.4~/~\textbf{97.4} & 93.0~/~\textbf{97.4} & 91.9~/~\textbf{96.5} & 94.9~/~\textbf{97.5} & 92.8~/~\textbf{97.2} \\
			& P-T-DI$^2$~--~(FGSM~/~FGS$^2$M)  & 99.0~/~\textbf{99.2} & \textbf{99.3}~/~99.1 & 99.2~/~\textbf{99.3} & 99.4~/~\textbf{99.5} & 99.2~/~\textbf{99.3} \\
			\hline
			\multirow{10}[4]{*}{\makecell[c]{AoE}} & I~--~(FGSM~/~FGS$^2$M)     & 94.7~/~\textbf{97.2} & 88.6~/~\textbf{93.3} & 92.5~/~\textbf{97.0} & 89.7~/~\textbf{93.1} & 91.4~/~\textbf{95.2} \\
			& MI~--~(FGSM~/~FGS$^2$M)    & 94.5~/~\textbf{96.6} & 90.1~/~\textbf{93.5} & 93.4~/~\textbf{97.0} & 90.5~/~\textbf{93.2} & 92.1~/~\textbf{95.1} \\
			& ~~DI$^2$~--~(FGSM~/~FGS$^2$M)   & 77.8~/~\textbf{89.1} & 76.3~/~\textbf{86.8} & 84.4~/~\textbf{93.7} & 77.8~/~\textbf{86.0} & 79.1~/~\textbf{88.9} \\
			& TI~--~(FGSM~/~FGS$^2$M)    & 94.0~/~\textbf{97.0} & 87.2~/~\textbf{92.3} & 92.5~/~\textbf{96.6} & 88.6~/~\textbf{92.3} & 90.6~/~\textbf{94.6} \\
			& Po~--~(FGSM~/~FGS$^2$M)    & 88.7~/~\textbf{92.8} & 82.4~/~\textbf{88.5} & 78.6~/~\textbf{85.9} & 87.1~/~\textbf{91.6} & 84.2~/~\textbf{89.7} \\
			& PI~--~(FGSM~/~FGS$^2$M)   & \textbf{98.1}~/~97.9 & 97.1~/~\textbf{97.4} & \textbf{98.1}~/~98.0 & 96.6~/~\textbf{96.9} & 97.5~/~\textbf{97.6} \\
			\cline{2-7}          
			& M-DI$^2$~--~(FGSM~/~FGS$^2$M) & 75.4~/~\textbf{84.8} & 74.6~/~\textbf{83.7} & 80.9~/~\textbf{89.8} & 76.6~/~\textbf{85.0} & 76.9~/~\textbf{85.8} \\
			& T-DI$^2$~--~(FGSM~/~FGS$^2$M) & 78.6~/~\textbf{89.1} & 75.9~/~\textbf{87.0} & 83.8~/~\textbf{93.6} & 76.8~/~\textbf{86.6} & 78.8~/~\textbf{89.1} \\
			& T-Po-M-DI$^2$~--~(FGSM~/~FGS$^2$M) & 79.4~/~\textbf{86.3} & 76.4~/~\textbf{83.8} & 77.0~/~\textbf{84.6} & 78.0~/~\textbf{84.6} & 77.7~/~\textbf{84.8} \\
			& P-T-DI$^2$~--~(FGSM~/~FGS$^2$M) & \textbf{94.8}~/~94.5 & \textbf{93.3}~/~93.1 & 97.0~/~\textbf{97.2} & 92.7~/~\textbf{93.5} & 94.5~/~\textbf{94.6} \\
			\hline
			\multirow{10}[4]{*}{\makecell[c]{Hold-out}} & I~--~(FGSM~/~FGS$^2$M)     & 1.2~/~\textbf{4.6} & 1.2~/~\textbf{3.4} & 0.0~/~\textbf{1.1} & 0.9~/~\textbf{1.9} & 0.8~/~\textbf{2.8} \\
			& MI~--~(FGSM~/~FGS$^2$M)    & 6.3~/~\textbf{6.5} & 3.6~/~\textbf{3.7} & \textbf{1.6}~/~1.4 & 3.0~/~\textbf{3.6} & 3.6~/~\textbf{3.8} \\
			& ~~DI$^2$~--~(FGSM~/~FGS$^2$M)   & 15.4~/~\textbf{28.8} & 13.8~/~\textbf{27.6} & 3.2~/~\textbf{8.6} & 9.4~/~\textbf{20.6} & 10.5~/~\textbf{21.4} \\
			& TI~--~(FGSM~/~FGS$^2$M)    & 1.6~/~\textbf{6.1} & 1.3~/~\textbf{4.2} & 0.3~/~\textbf{1.3} & 0.8~/~\textbf{3.2} & 1.0~/~\textbf{3.7} \\
			& Po~--~(FGSM~/~FGS$^2$M)    & 1.1~/~\textbf{4.8} & 0.9~/~\textbf{2.9} & 0.0~/~\textbf{0.4} & 0.3~/~\textbf{2.3} & 0.6~/~\textbf{2.6} \\
			& PI~--~(FGSM~/~FGS$^2$M)    & 22.4~/~\textbf{28.9} & 17.2~/~\textbf{23.2} & 4.2~/~\textbf{6.2} & 13.9~/~\textbf{20.9} & 14.4~/~\textbf{19.8} \\
			\cline{2-7}          
			& M-DI$^2$~--~(FGSM~/~FGS$^2$M) & 23.8~/~\textbf{31.5} & 24.1~/~\textbf{30.8} & 12.3~/~\textbf{14.2} & 21.3~/~\textbf{28.3} & 20.4~/~\textbf{26.2} \\
			& T-DI$^2$~--~(FGSM~/~FGS$^2$M) & 15.5~/~\textbf{29.8} & 15.9~/~\textbf{30.6} & 3.9~/~\textbf{9.7} & 11.3~/~\textbf{25.1} & 11.6~/~\textbf{23.8} \\
			& T-Po-M-DI$^2$~--~(FGSM~/~FGS$^2$M) & 34.7~/~\textbf{41.8} & 32.3~/~\textbf{40.4} & 17.3~/~\textbf{18.0} & 28.3~/~\textbf{34.4} & 28.2~/~\textbf{33.7} \\
			& P-T-DI$^2$~--~(FGSM~/~FGS$^2$M) & 46.9~/~\textbf{50.2} & 47.1~/~\textbf{50.8} & 14.2~/~\textbf{19.4} & 41.3~/~\textbf{44.7} & 37.4~/~\textbf{41.3} \\
			\hline
	\end{tabular}}%
	\label{tab:nt}%
\end{table*}%

\subsection{The Effect of Staircase Number $K$}
\label{exp:k}
In this section, we analyze the effect of the staircase number $K$ for the state-of-the-art FGSM-based attacks. Here we tune $K=2,4,8,...,256$. Please note that our methods only take $K\ge2$ as the input. $K=1$ denotes their corresponding FGSM-based baseline.

The experimental results of white-box attacks (Ensemble) are illustrated in Figure~\ref{fig:ensemble} (the discussion on AoE is left in Appendix Sec.2). A first glance shows that our FGS$^2$M variants have achieved an impressive improvement even when $K=2$. When the substitute model is an ensemble of six normally trained and two ensemble adversarial training models (\textit{i.e.} Figure~\ref{fig:ensemble}(c)), T-DI$^2$-FGS$^2$M significantly outperforms T-DI$^2$-FGSM by \textbf{17.1\%}. As the number of staircase increases, the success rate continues to rise and then remains stable after $K$ exceeds 64. For the methods whose success rates are already close to 100\%, \textit{e.g} Po-FGSM in Figure~\ref{fig:ensemble}(a), our FGS$^2$M variant does not degrade their great white-box performance.

We also depict the improvement curve for the black-box attacks (Hold-out) in Figure~\ref{fig:hold}. Compared with the vanilla FGSM implementation, our FGS$^2$M variants improve the transferability by a large margin as a whole. Specially, when transferring adversarial examples to normally trained models (Figure~\ref{fig:hold}(a)), DI-FGS$^2$M with $K=64$ sharply increases the success rate by \textbf{13.4\%} (from 15.4\% to 28.8\%).
Besides, as shown in Figure~\ref{fig:hold}(c), our methods also boost the attack performance on defenses, \textit{i.e.}, consistently outperform the corresponding baseline attacks by \textbf{4.3\% $\sim$ 16.5\%}.
Considering that the curves turn to remain stable when $K$ is big and most of methods reach the peak when $K=64$ in Figure~\ref{fig:ensemble} and Figure~\ref{fig:hold}, we set the staircase number $K=64$ in the following experiments. Note that the computational overhead of the percentage calculation is almost negligible compared to the cost of forward pass and backpropagation.

\begin{table*}[h]
	\centering
		\caption{The success rates (\%) of targeted FGSM-based/FGS$^2$M-based attacks. We study three models---Inc-v3$_{ens3}$, Inc-v3$_{ens4}$ and IncRes-v2$_{ens}$, and adversarial examples are crafted via an ensemble of eight of \{Inc-v3, Inc-v4, Res-152, Res-101, Res-50, IncRes-v2, Inc-v3$_{ens3}$, Inc-v3$_{ens4}$ and IncRes-v2$_{ens}$\}. In each column, ``-'' denote the hold-out model. 
		}
	\resizebox{0.75\linewidth}{!}{
		\begin{tabular}{c|c|c|c|c|c}
			\hline
			Metrics & Attacks & -Inc-v3$_{ens3}$ & -Inc-v3$_{ens4}$ & -IncRes-v2$_{ens}$ & AVG. \\
            \hline
            \hline
			\multirow{4}[2]{*}{\makecell[c]{\rotatebox{0}{Ensemble}}}
			& T-DI$^2$~--~(FGSM~/~FGS$^2$M) & 56.5~/~\textbf{78.1} & 56.5~/~\textbf{77.8} & 55.2~/~\textbf{76.7} & 56.1~/~\textbf{77.5} \\
			& T-M-DI$^2$~--~(FGSM~/~FGS$^2$M) & 44.8~/~\textbf{64.7} & 45.4~/~\textbf{64.9} & 48.9~/~\textbf{68.4} & 46.4~/~\textbf{66.0} \\
			& T-Po-M-DI$^2$~--~(FGSM~/~FGS$^2$M) & 58.9~/~\textbf{74.3} & 58.3~/~\textbf{75.3} & 60.6~/~\textbf{76.9} & 59.3~/~\textbf{75.5} \\
			& P-T-DI$^2$++~--~(FGSM~/~FGS$^2$M) & 94.0~/~\textbf{94.2} & \textbf{94.1}~/~93.8 & 94.2~/~\textbf{95.0} & 94.1~/~\textbf{94.3} \\
			\hline
			\multirow{4}[2]{*}{\makecell[c]{\rotatebox{0}{AoE}}} & T-DI$^2$~--~(FGSM~/~FGS$^2$M) & 43.4~/~\textbf{65.1} & 44.6~/~\textbf{65.5} & 46.2~/~\textbf{68.1} & 44.7~/~\textbf{66.2} \\
			& T-M-DI$^2$~--~(FGSM~/~FGS$^2$M) & 34.7~/~\textbf{52.1} & 36.2~/~\textbf{52.4} & 37.8~/~\textbf{54.5} & 36.2~/~\textbf{53.0} \\
			& T-Po-M-DI$^2$~--~(FGSM~/~FGS$^2$M) & 48.5~/~\textbf{63.8} & 48.9~/~\textbf{63.8} & 50.1~/~\textbf{65.5} & 49.2~/~\textbf{64.4} \\
			& P-T-DI$^2$++~--~(FGSM~/~FGS$^2$M) & 87.2~/~\textbf{87.3} & 87.0~/~\textbf{87.1} & \textbf{88.1}~/~87.9 & \textbf{87.4}~/~\textbf{87.4} \\
			\hline
			\multirow{4}[2]{*}{\makecell[c]{\rotatebox{0}{Hold-out}}} & T-DI$^2$~--~(FGSM~/~FGS$^2$M) & 13.4~/~\textbf{29.9} & 12.6~/~\textbf{30.0} & 10.4~/~\textbf{29.0} & 12.1~/~\textbf{29.6} \\
			& T-M-DI$^2$~--~(FGSM~/~FGS$^2$M) & 14.6~/~\textbf{25.5} & 14.5~/~\textbf{25.2} & 14.2~/~\textbf{24.3} & 14.4~/~\textbf{25.0} \\
			& T-Po-M-DI$^2$~--~(FGSM~/~FGS$^2$M) & 20.4~/~\textbf{33.9} & 20.0~/~\textbf{32.5} & 19.2~/~\textbf{30.8} & 19.9~/~\textbf{32.4} \\
			& P-T-DI$^2$++~--~(FGSM~/~FGS$^2$M) & 56.0~/~\textbf{60.3} & 56.5~/~\textbf{58.1} & 45.5~/~\textbf{51.7} & 52.7~/~\textbf{56.7} \\
			\hline
	\end{tabular}}%
	\label{tab:df}%
\end{table*}%

\begin{table*}[h]
\centering
\caption{The success rates (\%) of targeted FGSM-based/FGS$^2$M-based attacks. We study four models---HGD, NIPS-r3, R\&P and NRP, and adversarial examples are crafted via an ensemble of \{Inc-v3, Inc-v4, Res-152, Res-101, Res-50, IncRes-v2, Inc-v3$_{ens3}$, Inc-v3$_{ens4}$ and IncRes-v2$_{ens}$\}.
}
\resizebox{0.7\linewidth}{!}{
\begin{tabular}{c|c|c|c|c|c}
\hline
Attacks & HGD       & R\&P      & NIPS-r3   & NRP       & AVG.      \\
\hline
\hline
T-DI$^2$~--~(FGSM~/~FGS$^2$M)     & 23.2~/~\textbf{50.3} & 26.8~/~\textbf{52.9} & 24.8~/~\textbf{47.0} & 0.9~/~\textbf{3.9}   & 18.9~/~\textbf{38.5} \\
T-M-DI$^2$~--~(FGSM~/~FGS$^2$M)    & 23.3~/~\textbf{41.5} & 23.6~/~\textbf{41.0} & 22.4~/~\textbf{38.6} & 7.8~/~\textbf{13.4}  & 19.3~/~\textbf{33.6} \\
T-Po-M-DI$^2$~--~(FGSM~/~FGS$^2$M)  & 25.9~/~\textbf{41.6} & 26.1~/~\textbf{42.3} & 25.0~/~\textbf{38.2} & 10.6~/~\textbf{16.5} & 21.9~/~\textbf{34.7} \\
P-T-DI$^2$++~--~(FGSM~/~FGS$^2$M)  & 76.3~/~\textbf{78.4} & 82.3~/~\textbf{84.1} & 77.9~/~\textbf{80.1} & 28.8~/~\textbf{31.2} & 66.3~/~\textbf{68.5}
\\
\hline
\end{tabular}}
\label{4defense}
\end{table*}



\subsection{Attacking Normally Trained Models}
\label{exp:nt}
In this section, we compare ten FGSM-based attacks including I-FGSM, MI-FGSM, DI$^2$-FGSM, TI-FGSM, Po-FGSM, M-DI$^2$-FGSM, T-DI$^2$-FGSM, T-Po-M-DI$^2$-FGSM, P-T-DI$^2$-FGSM with our FGS$^2$M variants. In this experiment, four models including Inc-v3, Inc-v4, Res-152, and IncRes-v2 are considered. We select one model as the hold-out model to evaluate the transferability, and an ensemble of the rest three with the weight of each model $1/3$ serves as the substitute model. 

As indicated in Table~\ref{tab:nt}, our proposed FGS$^2$M variants effectively boost both the white-box and black-box attacks. On average, they increase the success rate in Ensemble, AoE and Hold-out cases by \textbf{2.1\%}, \textbf{5.3\%} and \textbf{5.1\%}, respectively. This demonstrates that our adversarial examples are more close to the \textit{global optimal  attack region}. 

From the results of Table~\ref{tab:nt}, we also observe that several methods, especially for these integrated with diversity input patterns (DI$^2$), suffer from SM which cannot well utilize the gradient with respect to the random input transformation. 
Specifically, DI$^2$-FGSM only successfully attack 93.6\% images against the substitute model (Ensemble).
The average success rate of each white-box model (AoE) is even reduced to 79.1\%, 
and merely 10.5\% images transfer to the black-box model (Hold-out) on average. With the help of our S$^2$M at each iteration, we dramatically alleviate the poor gradient estimation problem, that is, increasing the success rate in Ensemble and AoE cases by \textbf{4.8\%} and \textbf{9.8\%}, respectively. Furthermore, in the Hold-out case our DI$^2$-FGS$^2$M remarkably improves the transferability by \textbf{10.9\%}.

Another observation from the results is that the staircase sign perturbation seems to be less effective on vanilla MI but more effective for other momentum-based methods in the black-box manner, \textit{e.g.}, M-DI$^2$. It may be because that vanilla MI will impose a more significant noise curing~\cite{pom}, thus causing a lack of diversity and adaptability of update direction derived by our FGS$^2$M variants. 


\subsection{Attacking Black-box Robust Defenses}
\label{exp:df}
Defense models are shown to effectively withstand the transferable adversarial examples in the black-box scenario. Therefore, here we compare four stronger FGSM-based attacks including T-DI$^2$-FGSM, T-M-DI$^2$-FGSM, T-Po-M-DI$^2$-FGSM, P-T-DI$^2$-FGSM++ with our FGS$^2$M variants, and adversarial examples are crafted via an ensemble of part or all of \{Inc-v3, Inc-v4, Res-152, Res-101, Res-50, IncRes-v2, Inc-v3$_{ens3}$, Inc-v3$_{ens4}$ and IncRes-v2$_{ens}$\}.

As demonstrated in Table~\ref{tab:df}, regardless of the attacks are white-box or black-box, our methods generally surpass the vanilla FGSM-based methods. Specifically, in the white-box manner, FGS$^2$M-based attacks, on average, outperform FGSM-based ones by \textbf{14.4\%} (Ensemble) and \textbf{13.4\%} (AoE). This again demonstrates that our method can effectively alleviate the poor gradient estimation problem caused by SM. Besides, even under the more challenging black-box attack manner (shown in Table~\ref{tab:df} and Table~\ref{4defense}), our proposed attacks can significantly improve the transferability by \textbf{12.8\%} on average.
Remarkably, compared with T-DI$^2$-FGSM, which only successfully transfers 23.2\% adversarial examples to HGD, our T-DI$^2$-FGS$^2$M can further enhance the transferability by \textbf{27.1\%}. Besides, when crafting adversarial perturbation by our proposed FGS$^2$M implementation of P-T-DI$^2$++, resultant adversarial examples can fool R\&P with a high success rate of \textbf{84.1\%}.  



\begin{table*}[t]
\centering
\caption{The white-box success rate (\%) of targeted FGSM-based/FGS$^2$M-based attacks. We study three models---Res152$_B$, Res152$_D$ and ResNext101$_{DA}$, and adversarial examples are crafted via each of them. Following~\cite{pifgsm}, here we set $\beta=2.5$ and $\gamma=0.25\beta\alpha$ for PI.
}
\resizebox{0.6\linewidth}{!}{
\begin{tabular}{c|c|c|c|c}
\hline
Attacks & Res152$_B$  & Res152$_D$         & ResNext101$_{DA}$        & AVG.      \\
\hline
\hline
I~--~(FGSM~/~FGS$^2$M)       & 11.2~/~\textbf{20.2} & 9.5~/~\textbf{18.2}  & 9.6~/~\textbf{21.4}  & 10.1~/~\textbf{19.9} \\
MI~--~(FGSM~/~FGS$^2$M)      & 6.7~/~\textbf{12.8}  & 5.3~/~\textbf{11.4}  & 5.2~/~\textbf{11.7}  & 5.7~/~\textbf{12.0}  \\
DI~--~(FGSM~/~FGS$^2$M)      & 0.7~/~\textbf{5.7}   & 1.0~/~\textbf{4.2}   & 0.8~/~\textbf{5.4}   & 0.8~/~\textbf{5.1}   \\
TI~--~(FGSM~/~FGS$^2$M)      & 6.2~/~\textbf{11.0}  & 5.2~/~\textbf{8.6}   & 5.5~/~\textbf{9.3}   & 5.6~/~\textbf{9.6}   \\
Po~--~(FGSM~/~FGS$^2$M)      & 8.5~/~\textbf{15.1}  & 7.3~/~\textbf{13.2}  & 7.3~/~\textbf{15.9}  & 7.7~/~\textbf{14.7}  \\
PI~--~(FGSM~/~FGS$^2$M)      & 11.8~/~\textbf{20.0} & 10.5~/~\textbf{17.8} & 10.5~/~\textbf{20.5} & 10.9~/~\textbf{19.4} \\
\hline
\end{tabular}}
\label{white-box-df}
\end{table*}
\begin{table*}[t]
\caption{The success rates (\%) of targeted FGSM-based/FGS$^2$M-based attacks under $\epsilon=4,\,8,\,12$. Adversarial examples are crafted via an ensemble of \{Inc-v3, Inc-v4, Res-152, Res-101, Res-50, IncRes-v2, Inc-v3$_{ens4}$ and IncRes-v2$_{ens}$\} and the hold-out model is Inc-v3$_{ens3}$.}
\centering
\resizebox{0.75\linewidth}{!}{
\begin{tabular}{c|c|p{2cm}<{\centering}|p{2cm}<{\centering}|p{2cm}<{\centering}}
\hline
Perturbations & Attacks   &  Ensemble& AoE      &  Hold-out \\
\hline
\hline
\multirow{4}[1]{*}{$\epsilon=4$} &
T-DI$^2$~--~(FGSM~/~FGS$^2$M)      &   1.6~/~\textbf{5.5} & 2.5~/~\textbf{6.7}   &  0.1~/~\textbf{0.3} \\
&T-M-DI$^2$~--~(FGSM~/~FGS$^2$M)    &  1.3~/~\textbf{2.9} & 1.7~/~\textbf{3.6}   &  0.2~/~\textbf{0.4}  \\
&T-Po-M-DI$^2$~--~(FGSM~/~FGS$^2$M) & 1.9~/~\textbf{4.3}  & 3.3~/~\textbf{6.7}   & 0.2~/~\textbf{0.4}   \\
&P-T-DI$^2$++~--~(FGSM~/~FGS$^2$M)  & 19.4~/~\textbf{21.3} & 21.0~/~\textbf{21.6} &  2.1~/~\textbf{3.6} \\
\hline
\multirow{4}[1]{*}{$\epsilon=8$} &
T-DI$^2$~--~(FGSM~/~FGS$^2$M)      &  15.2~/~\textbf{34.5}  & 14.1~/~\textbf{29.5} &  1.5~/~\textbf{5.9}\\
&T-M-DI$^2$~--~(FGSM~/~FGS$^2$M)    & 11.5~/~\textbf{21.8}   & 10.3~/~\textbf{19.3} &  2.6~/~\textbf{4.8}\\
&T-Po-M-DI$^2$~--~(FGSM~/~FGS$^2$M) &  15.5~/~\textbf{28.1}  & 17.1~/~\textbf{27.5} &  3.0~/~\textbf{6.6}\\
&P-T-DI$^2$++~--~(FGSM~/~FGS$^2$M)  &  65.2~/~\textbf{66.1}& 56.9~/~\textbf{57.4} &  17.1~/~\textbf{20.1}\\
\hline
\multirow{4}[1]{*}{$\epsilon=12$} &
T-DI$^2$~--~(FGSM~/~FGS$^2$M)      & 35.4~/~\textbf{60.0} & 30.2~/~\textbf{50.8} &  6.1~/~\textbf{18.6}\\
&T-M-DI$^2$~--~(FGSM~/~FGS$^2$M)    &  27.9~/~\textbf{45.3} & 23.1~/~\textbf{38.1} &  7.6~/~\textbf{14.9} \\
&T-Po-M-DI$^2$~--~(FGSM~/~FGS$^2$M) &  34.9~/~\textbf{52.1}  & 32.0~/~\textbf{46.8} & 10.1~/~\textbf{17.5} \\
&P-T-DI$^2$++~--~(FGSM~/~FGS$^2$M)  & 84.9~/~\textbf{86.4}& 76.7~/~\textbf{77.0} &38.6~/~\textbf{41.9}  \\
\hline
\end{tabular}}
\label{eps4}
\end{table*}
\subsection{Attacking White-box Robust Defenses}
\label{exp:wdf}
To further demonstrate the superiority of our proposed method, we consider three feature denoising models (Res152$_B$, Res152$_D$, ResNext101$_{DA}$)~\cite{featuredenoising} which are even robust against white-box attacks. Since Gao \textit{et al.}~\cite{pifgsm} have shown that transferable techniques are less effective for attacking these feature denoising model, here we do not consider the combined version of these attacks and just 
report white-box results of I, MI, DI, TI, Po and PI.

As shown in Table~\ref{white-box-df}, our FGS$^2$M variants consistently outperform vanilla FGSM ones in this challenging white-box scenario.  Compared with vanilla FGSM implementation of I, MI, DI, TI, Po and PI, our FGS$^2$M variants can surpass them by 9.8\%, 6.3\%, 4.3\%, 4.0\%, 7.0\% and 8.5\%, respectively. This result again demonstrates that our method can yield better adversarial solution for attacking 
defense models.

\subsection{Experiments for Different $\epsilon$}
\label{exp:epsilon}
In Table~\ref{eps4}, we discuss the attack success rates with respective to maximum perturbation $\epsilon$, \textit{i.e.}, 4, 8 and 12. 
Similar with the result of $\epsilon=16$, our proposed methods consistently outperform the vanilla FGSM-based methods. Besides, as $\epsilon$ grows, the gap between our FGS$^2$M variants and FGSM baselines can be further enlarged.

\subsection{Insight into Staircase Sign Method}
\label{insight}
\begin{figure}[t]
    \centering
    \includegraphics[height=5.25cm]{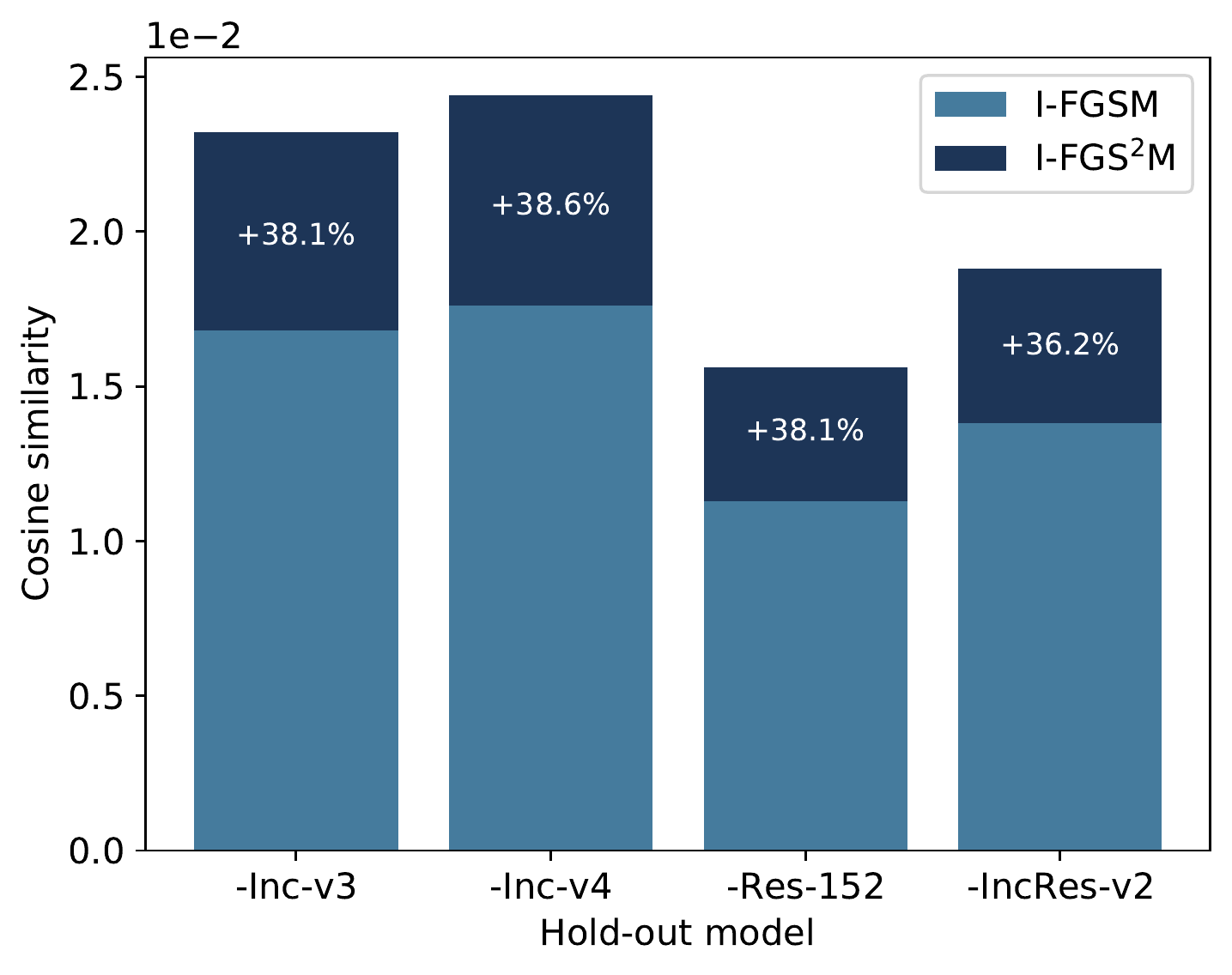}
	\caption{We compared I-FGSM with our I-FGS$^2$M in terms of cosines similarity between update direction on the substitute model and raw gradient on the victim model. Adversarial examples are crafted via an ensemble of three of \{Inc-v3, Inc-v4, Res-152 and IncRes-v2\}, and ``-" denotes the hold-out model.}
	\label{fgm}
\end{figure}

To better understand the effect of our staircase sign design, in this section, we give an insight into it from a perspective of resultant update direction. As described in Sec.~\ref{methodology},
the motivation behind S$^2$M is to mitigate the problem of poor gradient estimation resulting from SM. To prove our proposed method does have this advantage, here we investigate the cosine similarity between raw gradient on the victim model and update direction on the substitute model. The result is shown in Figure~\ref{fgm}. Compared with widely used SM, our proposed S$^2$M can boost cosine similarity by an additional \textbf{37.8\%} on average. This result demonstrates that S$^2$M can significantly narrow the gap between the substitute model and the victim model.

\section{Conclusion}
In this paper, we rethink the limitation of Sign Method (SM) applied by state-of-the-art FGSM family, and  empirically and experimentally demonstrate that it causes poor gradient estimation. To address this issue, we propose a simple but effective Staircase Sign Method (S$^2$M) to boost transferability. With the help of staircase weights, our methods effectively fool both white-box models and black-box models. 
Extensive experiments on the ImageNet dataset demonstrate the effectiveness of our FGS$^2$M-based attacks, which significantly improves the transferability by \textbf{5.1\%} for normally trained models and \textbf{12.8\%} for adversarially trained defenses on average. 

{\small
\bibliographystyle{ieee_fullname}
\bibliography{egbib}
}
\clearpage
\appendix
\section{Setup}
\label{app:setup}

\begin{figure*}[t]
	\centering
	\includegraphics[height=5cm]{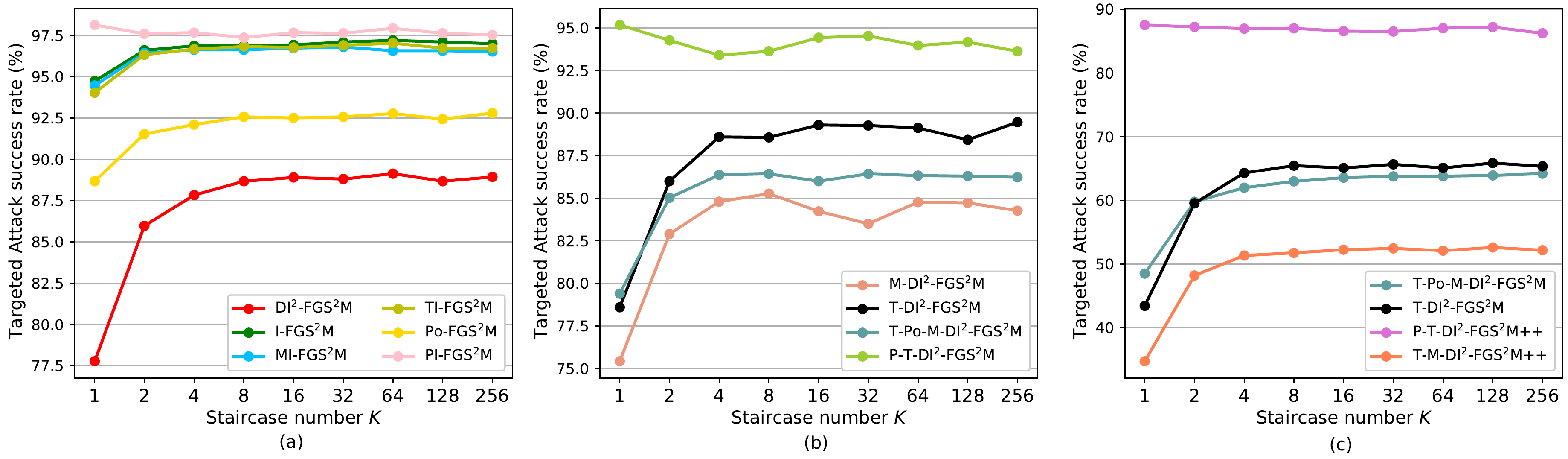}
	\caption{The success rates (\%) of targeted white-box attack (AoE) for different methods w.r.t. staircase number $K$ ($K\!=\!1$ denotes SM and $K\!\ge\!2$ denotes S$^2$M). 
		For (a) and (b), the adversarial examples are crafted via an ensemble of Inc-v4, IncRes-v2 and Res-152, and the hold-out model is Inc-v3. For (c), the white-box models are an ensemble of Inc-v3, Inc-v4, IncRes-v2, Res-152, Res-101, Res-50, Inc-v3$_{ens4}$ and IncRes-v2$_{ens}$, and the hold-out model is Inc-v3$_{ens3}$.}
	\label{fig:aoe}
\end{figure*}
\textbf{Parameters}: For targeted attacks, we adopt the same parameter setting in our paper. For non-targeted attacks, here we following the previous works~\cite{pifgsm,sinifgsm}. In our experiments, the maximum perturbation $\epsilon$ is set to 16 unless otherwise stated0. The iteration $T$ of all methods is 10, and thus the step size $\alpha=\epsilon/T=1.6$.
For MI-FGSM~\cite{mifgsm}, the decay factor $\mu=1.0$. For DI-FGSM~\cite{difgsm}, the transformation probability $p=0.7$. For TI-FGSM~\cite{tifgsm}, the Gaussian kernel length is $15\times15$. For PI-FGSM~\cite{pifgsm}, the amplification factor $\beta=5$, the project factor $\gamma=\beta\alpha$, and the project kernel length is $3\times 3$. For SI-FGSM~\cite{sinifgsm}, the number of scale copies $m=5$.
For our S$^2$M, the number of staircase $K$ is set to 64. Please note that the parameters of each method are fixed no matter what methods are combined.

\section{The Effect of Staircase Number $K$}
\label{appendix:k}

Here we show the experimental results of white-box attacks, \textit{i.e.}, Figure~\ref{fig:aoe} for AoE. In this section, we analyze the effect of the staircase number $K$ for state-of-the-art FGSM-based attacks. Here we tune $K=2,4,8,...,256$. Similar to the observation in Sec.~\textcolor{red}{4.2}, our FGS$^2$M variants can also improve the success rates by a large margin even when $K=2$ and the success rate continues to increase and then remain stable after $K$ exceeds 64.

\section{Experiments for Non-targeted Attacks}
\label{appendix:non-targeted}
Due to space limitation, we mainly discuss the more challenging targeted attacks in our paper. Since the poor gradient estimation problem is caused by SM, crafting adversarial perturbations by our proposed S$^2$M can also boost non-targeted attacks. In this section, we report our experimental results to demonstrate the effectiveness of our methods.

Specifically, we compare six FGSM-based attacks, including I-FGSM, MI-FGSM, DI$^2$-FGSM, TI-FGSM, SI-FGSM and PI-FGSM, with our FGS$^2$M variants. In this experiment, we study nine models inlcuding Inc-v3, Inc-v4, IncRes-v2, Res-152, Res-101, Res-50, Inc-v3$_{ens3}$, Inc-v3$_{ens4}$ and IncRes-v2$_{ens}$. Since non-targeted attacks are less challenging than targeted attacks, we craft adversarial examples via one model instead of an ensemble of models. 

As demonstrated in Table~\ref{tab:nt_inc-v3}, Table~\ref{tab:nt_inc-v4}, Table~\ref{tab:nt_incres} and Table~\ref{tab:nt_res}, our FGS$^2$M-based attacks consistently outperform FGSM-based ones in both the white-box and black-box manner. For the black-box manner, we significantly improve the transferability by \textbf{8.2\%} on average.
Remarkably, when adversarial examples are crafted via IncRes-v2 by SI-FGS$^2$M, we can transfer an extra \textbf{19.1\%} adversarial examples to Inc-v3$_{ens3}$. For the white-box manner, our FGS$^2$M variants can increase the success rate of white-box attacks toward 100\%. As demonstrated in Table~\ref{tab:nt_inc-v3}, I-FGSM only successfully attacks Inc-v3 with a 99.2\% success rate. But with the help of our staircase weights, our I-FGS$^2$M can achieve an success rate of \textbf{100\%}. 

\begin{table*}[h]
	\caption{The success rates (\%) of non-targeted FGSM-based/FGS$^2$M-based attacks w.r.t. adversarial examples crafted via Inc-v3. We study nine models---Inc-v3, Inc-v4, IncRes-v2, Res-152, Res-101, Res-50, Inc-v3$_{ens3}$, Inc-v3$_{ens4}$ and IncRes-v2$_{ens}$ here.}
	\centering
	\resizebox{1.0\linewidth}{!}{
		\begin{tabular}{c|c|c|c|c|c|c|c|c|c|c|c}
			\hline
			& Attacks & Inc-v3* & Inc-v4 & IncRes-v2 & Res-152 & Res-101 & Res-50 & Inc-v3$_{ens3}$  & Inc-v3$_{ens4}$  & IncRes-v2$_{ens}$ & AVG. \\
			\hline
			\hline
			\multirow{6}{*}{\makecell[c]{Inc-v3}} 
			& I     & 99.2~/~\textbf{100.0} & 30.0~/~\textbf{40.6} & 21.5~/~\textbf{34.9} & 18.9~/~\textbf{26.6} & 20.8~/~\textbf{29.6} & 23.3~/~\textbf{32.6} & 12.1~/~\textbf{16.0} & 12.1~/~\textbf{17.3} & 4.9~/~\textbf{8.4} & 18.0~/~\textbf{25.8} \\
			
			& MI    & 99.2~/~\textbf{100.0} & 55.7~/~\textbf{57.5} & 51.2~/~\textbf{55.1} & \textbf{44.0}~/~\textbf{44.0} & 44.6~/~\textbf{46.9} & 49.9~/~\textbf{51.8} & 21.9~/~\textbf{22.7} & 20.7~/~\textbf{23.1} & 11.0~/~\textbf{11.6} & 37.4~/~\textbf{39.1} \\
			
			& DI    & 99.9~/~\textbf{100.0} & 52.5~/~\textbf{67.0} & 42.5~/~\textbf{57.8} & 32.4~/~\textbf{42.9} & 36.0~/~\textbf{48.4} & 41.4~/~\textbf{52.4} & 13.9~/~\textbf{22.0} & 14.6~/~\textbf{22.7} & 6.9~/~\textbf{11.6} & 30.0~/~\textbf{40.6} \\
			
			& TI    & 99.1~/~\textbf{100.0} & 27.5~/~\textbf{35.9} & 14.0~/~\textbf{24.5} & 16.3~/~\textbf{23.0} & 17.9~/~\textbf{25.6} & 22.1~/~\textbf{28.1} & 17.8~/~\textbf{28.0} & 16.5~/~\textbf{26.6} & 10.4~/~\textbf{16.9} & 17.8~/~\textbf{26.1} \\
			
			& SI    & \textbf{100.0}~/~\textbf{100.0} & 53.8~/~\textbf{69.3} & 47.2~/~\textbf{64.9} & 39.4~/~\textbf{53.5} & 45.3~/~\textbf{58.9} & 48.7~/~\textbf{61.3} & 21.7~/~\textbf{33.4} & 22.6~/~\textbf{37.2} & 10.8~/~\textbf{20.1} & 36.2~/~\textbf{49.8} \\
			
			& PI    & \textbf{100.0}~/~\textbf{100.0} & 54.5~/~\textbf{62.9} & 47.4~/~\textbf{55.9} & 39.7~/~\textbf{47.6} & 43.0~/~\textbf{48.7} & 48.2~/~\textbf{52.1} & 26.3~/~\textbf{31.3} & 25.4~/~\textbf{29.0} & 15.5~/~\textbf{19.2} & 37.5~/~\textbf{43.3} \\
			\hline
	\end{tabular}}%
	\label{tab:nt_inc-v3}%
\end{table*}%

\begin{table*}[h]
	\caption{The success rates (\%) of non-targeted FGSM-based/FGS$^2$M-based attacks w.r.t. adversarial examples crafted via Inc-v4. We study nine models---Inc-v3, Inc-v4, IncRes-v2, Res-152, Res-101, Res-50, Inc-v3$_{ens3}$, Inc-v3$_{ens4}$ and IncRes-v2$_{ens}$ here.}
	\centering
	\resizebox{1.0\linewidth}{!}{
		\begin{tabular}{c|c|c|c|c|c|c|c|c|c|c|c}
			\hline
			& Attacks & Inc-v3 & Inc-v4* & IncRes-v2 & Res-152 & Res-101 & Res-50 & Inc-v3$_{ens3}$  & Inc-v3$_{ens4}$  & IncRes-v2$_{ens}$ & AVG. \\
			\hline
			\hline
			\multirow{6}{*}{\makecell[c]{Inc-v4}} 
			& I     & 43.2~/~\textbf{56.9} & 99.2~/~\textbf{100.0} & 26.3~/~\textbf{39.2} & 25.2~/~\textbf{34.6} & 25.9~/~\textbf{36.1} & 30.9~/~\textbf{39.8} & 12.0~/~\textbf{16.9} & 12.6~/~\textbf{19.0 }& 6.4~/~\textbf{10.5} & 22.8~/~\textbf{31.6} \\
			        
			& MI    & 70.8~/~\textbf{73.1} & 99.2~/~\textbf{100.0} & 58.1~/~\textbf{60.4} & 52.2~/~\textbf{53.5} & 53.8~/~\textbf{54.8} & 56.6~/~\textbf{59.1} & 23.8~/~\textbf{26.0} & 23.8~/~\textbf{25.0} & 12.7~/~\textbf{13.9} & 44.0~/~\textbf{45.7} \\
			        
			& DI    & 64.5~/~\textbf{75.3} & 99.1~/~\textbf{100.0} & 48.3~/~\textbf{63.5} & 38.6~/~\textbf{48.5} & 40.0~/~\textbf{50.4} & 44.3~/~\textbf{54.5} & 16.0~/~\textbf{21.5} & 16.3~/~\textbf{22.9} & 8.6~/~\textbf{13.7} & 34.6~/~\textbf{43.8} \\
			         
			& TI    & 36.8~/~\textbf{46.6} & 99.2~/~\textbf{100.0} & 16.8~/~\textbf{27.9} & 20.8~/~\textbf{27.9} & 18.9~/~\textbf{26.9} & 22.3~/~\textbf{31.9} & 16.2~/~\textbf{26.8} & 19.8~/~\textbf{27.7}& 11.6~/~\textbf{17.9} & 20.4~/~\textbf{29.2} \\
			          
			& SI    & 72.0~/~\textbf{81.7} & \textbf{100.0}~/~\textbf{100.0} & 57.0~/~\textbf{71.7} & 51.1~/~\textbf{62.3} & 52.1~/~\textbf{64.2} & 56.7~/~\textbf{68.6} & 26.6~/~\textbf{43.7} & 28.0~/~\textbf{45.4} & 16.9~/~\textbf{29.8} & 45.1~/~\textbf{58.4} \\
			          
			& PI    & 68.7~/~\textbf{74.9} & \textbf{100.0}~/~\textbf{100.0} & 51.4~/~\textbf{60.2} & 45.4~/~\textbf{53.3} & 44.5~/~\textbf{52.8} & 52.2~/~\textbf{57.7} & 28.2~/~\textbf{35.4} & 27.8~/~\textbf{33.6} & 19.7~/~\textbf{23.6} & 42.2~/~\textbf{48.9} \\
			\hline
	\end{tabular}}%
	\label{tab:nt_inc-v4}%
\end{table*}%

\begin{table*}[h]
	\caption{The success rates (\%) of non-targeted FGSM-based/FGS$^2$M-based attacks w.r.t. adversarial examples crafted via IncRes-v2. We study nine models---Inc-v3, Inc-v4, IncRes-v2, Res-152, Res-101, Res-50, Inc-v3$_{ens3}$, Inc-v3$_{ens4}$ and IncRes-v2$_{ens}$ here.}
	\centering
	\resizebox{1.0\linewidth}{!}{
		\begin{tabular}{c|c|c|c|c|c|c|c|c|c|c|c}
			\hline
			& Attacks & Inc-v3 & Inc-v4 & IncRes-v2* & Res-152 & Res-101 & Res-50 & Inc-v3$_{ens3}$  & Inc-v3$_{ens4}$  & IncRes-v2$_{ens}$ & AVG. \\
			\hline
			\hline
			\multirow{6}{*}{\makecell[c]{IncRes-v2}} 
			& I     & 46.7~/~\textbf{59.5} & 38.2~/~\textbf{49.0} & 99.2~/~\textbf{100.0} & 25.4~/~\textbf{36.5} & 28.2~/~\textbf{39.9} & 30.7~/~\textbf{42.1} & 13.2~/~\textbf{21.4} & 13.0~/~\textbf{19.5} & 8.3~/~\textbf{15.2} & 25.5~/~\textbf{35.4} \\
			          
			& MI    & \textbf{76.1}~/~75.8 & 67.9~/~\textbf{68.8} & 99.2~/~\textbf{100.0} & \textbf{57.6}~/~56.3 & 57.9~/~\textbf{58.9} & 61.3~/~\textbf{63.3} & 32.0~/~\textbf{34.7} & 28.4~/~\textbf{28.8} & 20.5~/~\textbf{22.0} & 50.2~/~\textbf{51.1} \\
			        
			& DI    & 71.4~/~\textbf{79.5} & 65.3~/~\textbf{76.6} & 98.5~/~\textbf{99.7} & 47.8~/~\textbf{58.3} & 49.6~/~\textbf{59.8} & 54.38~/~\textbf{64.7} & 19.5~/~\textbf{31.0} & 19.1~/~\textbf{28.1} & 12.2~/~\textbf{22.6} & 42.5~/~\textbf{52.6} \\
			          
			& TI    & 43.5~/~\textbf{52.2} & 41.2~/~\textbf{47.8} & 98.8~/~\textbf{99.9} & 26.3~/~\textbf{31.4} & 28.6~/~\textbf{34.5} & 30.2~/~\textbf{38.0} & 26.7~/~\textbf{35.6} & 24.7~/~\textbf{36.4} & 20.8~/~\textbf{32.4} & 30.3~/~\textbf{38.5} \\
			        
			& SI    & 74.0~/~\textbf{83.9} & 64.2~/~\textbf{75.7} & \textbf{99.9}~/~\textbf{99.9} & 51.7~/~\textbf{66.1} & 52.9~/~\textbf{66.5} & 60.5~/~\textbf{73.3} & 29.6~/~\textbf{48.7} & 28.8~/~\textbf{44.3} & 22.1~/~\textbf{40.7} & 48.0~/~\textbf{62.4} \\
			         
			& PI    & 72.6~/~\textbf{79.0} & 64.1~/~\textbf{72.8} & \textbf{100.0}~/~\textbf{100.0} & 52.2~/~\textbf{59.0} & 53.8~/~\textbf{61.2} & 56.9~/~\textbf{63.9} & 34.3~/~\textbf{43.4} & 30.9~/~\textbf{38.5} & 25.6~/~\textbf{33.2} & 48.8~/~\textbf{56.4} \\
			\hline
	\end{tabular}}%
	\label{tab:nt_incres}%
\end{table*}%

\begin{table*}[h]
	\caption{The success rates (\%) of non-targeted FGSM-based/FGS$^2$M-based attacks w.r.t. adversarial examples crafted via Res-152. We study nine models---Inc-v3, Inc-v4, IncRes-v2, Res-152, Res-101, Res-50, Inc-v3$_{ens3}$, Inc-v3$_{ens4}$ and IncRes-v2$_{ens}$ here.}
	\centering
	\resizebox{1.0\linewidth}{!}{
		\begin{tabular}{c|c|c|c|c|c|c|c|c|c|c|c}
			\hline
			& Attacks & Inc-v3 & Inc-v4 & IncRes-v2 & Res-152* & Res-101 & Res-50 & Inc-v3$_{ens3}$  & Inc-v3$_{ens4}$  & IncRes-v2$_{ens}$ & AVG. \\
			\hline
			\hline
			\multirow{6}{*}{\makecell[c]{IncRes-v2}} 
			& I     & 31.3~/~\textbf{43.8} & 25.9~/~\textbf{35.6} & 17.7~/~\textbf{31.6} & 98.7~/~\textbf{99.5} & 67.3~/~\textbf{80.8} & 66.1~/~\textbf{78.0} & 12.2~/~\textbf{17.7} & 13.3~/~\textbf{19.4} & 7.6~/~\textbf{12.6} & 30.2~/~\textbf{39.9} \\
			          
			& MI    & 55.9~/~\textbf{59.5} & 50.0~/~\textbf{52.2} & 45.9~/~\textbf{50.3} & 98.7~/~\textbf{99.5} & 85.2~/~\textbf{88.0} & 83.3~/~\textbf{87.6} & 26.9~/~\textbf{29.7} & 25.7~/~\textbf{26.8} & 15.3~/~\textbf{16.4} & 48.5~/~\textbf{51.3} \\
			          
			& DI    & 60.6~/~\textbf{74.0} & 56.5~/~\textbf{68.2} & 51.0~/~\textbf{65.7} & 98.4~/~\textbf{99.6} & 86.8~/~\textbf{93.6} & 84.2~/~\textbf{92.0} & 21.2~/~\textbf{33.6} & 20.1~/~\textbf{31.9} & 13.0~/~\textbf{21.6} & 49.2~/~\textbf{60.1} \\
			         
			& TI    & 25.5~/~\textbf{32.6} & 21.9~/~\textbf{28.2} & 11.0~/~\textbf{19.1} & 98.2~/~\textbf{99.2} & 53.3~/~\textbf{62.8} & 48.2~/~\textbf{56.3} & 18.4~/~\textbf{26.2} & 18.7~/~\textbf{25.9} & 12.6~/~\textbf{20.0} & 26.2~/~\textbf{33.9} \\
			         
			& SI    & 43.6~/~\textbf{56.5} & 40.3~/~\textbf{50.5} & 32.4~/~\textbf{47.1} & \textbf{99.8}~/~\textbf{99.8} & 84.7~/~\textbf{91.6} & 83.7~/~\textbf{89.9} & 19.0~/~\textbf{33.0} & 18.9~/~\textbf{31.4} & 12.5~/~\textbf{22.4} & 41.9~/~\textbf{52.8} \\
			          
			& PI    & 57.5~/~\textbf{63.9} & 50.3~/~\textbf{57.8} & 47.4~/~\textbf{55.0} & 99.6~/~\textbf{99.7} & 82.7~/~\textbf{90.6} & 81.8~/~\textbf{87.9} & 31.7~/~\textbf{38.1} & 29.5~/~\textbf{37.4} & 21.2~/~\textbf{27.1} & 50.3~/~\textbf{57.2} \\
			\hline
	\end{tabular}}%
	\label{tab:nt_res}%
\end{table*}%

\end{document}